\documentclass[11pt]{article}

\usepackage[final]{acl}

\usepackage{times}
\usepackage{latexsym}
\usepackage{graphicx}

\usepackage[T1]{fontenc}
\usepackage[utf8]{inputenc}

\usepackage{microtype}
\usepackage{amsmath}

\usepackage{inconsolata}

\usepackage{graphicx}

\usepackage{booktabs}
\usepackage{xcolor}
\usepackage{colortbl}
\usepackage{multirow}
\usepackage{tikz}
\usepackage{pgfplots}
\pgfplotsset{compat=1.16}

\definecolor{lowperf}{RGB}{255,245,240}
\definecolor{midperf}{RGB}{252,146,114}
\definecolor{highperf}{RGB}{200,0,0}
\definecolor{bestperf}{RGB}{127,0,0}

\definecolor{lowgreen}{RGB}{247,252,245}
\definecolor{midgreen}{RGB}{161,217,155}
\definecolor{highgreen}{RGB}{49,163,84}
\definecolor{bestgreen}{RGB}{0,109,44}

\definecolor{perf0}{RGB}{255,255,255}      % White for lowest
\definecolor{perf10}{RGB}{252,255,250}     % Extremely light green
\definecolor{perf20}{RGB}{240,252,236}     % Very light green  
\definecolor{perf30}{RGB}{220,245,210}     % Light green
\definecolor{perf40}{RGB}{190,235,180}     % Light-medium green
\definecolor{perf50}{RGB}{160,225,150}     % Medium green
\definecolor{perf60}{RGB}{130,210,120}     % Medium-dark green
\definecolor{perf70}{RGB}{100,195,90}      % Dark green
\definecolor{perf80}{RGB}{70,180,60}       % Darker green
\definecolor{perf90}{RGB}{40,165,30}       % Darkest green

\title{Binary Token-Level Classification with DeBERTa for All-Type MWE Identification: A Lightweight Approach with Linguistic Enhancement}

\author{Diego Rossini \and  Lonneke van der Plas\\
  Università della Svizzera Italiana\\
  \texttt{diego.rossini@usi.ch \and lonneke.vanderplas@usi.ch} \\}

\begin{document}
\maketitle
\begin{abstract}
We present a comprehensive approach for multiword expression (MWE) identification that combines binary token-level classification, linguistic feature integration, and data augmentation.\footnote{Model available at \url{https://huggingface.co/DiegoRossini/mwe-detection-deberta}. Code at \url{https://github.com/DiegoRossini/binary-mwe-detection}.} Our DeBERTa-v3-large model achieves 69.8\% F1 on the CoAM dataset, surpassing the best results (Qwen-72B, 57.8\% F1) on this dataset by 12 points while using 165$\times$ fewer parameters. We achieve this performance by \textbf{(1)} reformulating detection as binary token-level START/END/INSIDE classification rather than span-based prediction, \textbf{(2)} incorporating NP chunking and dependency features that help discontinuous and NOUN-type MWEs identification, and \textbf{(3)} applying oversampling that addresses severe class imbalance in the training data. We confirm the generalization of our method on the STREUSLE dataset, achieving 78.9\% F1. These results demonstrate that carefully designed smaller models can substantially outperform LLMs on structured NLP tasks, with important implications for resource-constrained deployments.
\end{abstract}

\section{Introduction}

Multiword expressions (MWEs) challenge NLP systems due to their syntactic flexibility and semantic non-compositionality \citep{constant2017survey, sag2002}. They are also widely acknowledged to be frequent in natural language \citep{baldwin2010}. Accurate detection nevertheless remains difficult, particularly for discontinuous patterns in which components may be separated by intervening material, as in \textit{{\underline{look} the information} \underline{up}}, where \textit{look...up} forms the MWE while \textit{the information} does not \citep{rohanian2019}.

MWE identification is traditionally formulated as sequence-labeling tasks using BIO \citep{ramshaw1995} or extended tagsets (e.g., BILOU \citep{ratinov2009}) to capture discontinuities. These multi-class labeling schemes assign each token a single label from a predefined tagset. Current state-of-the-art approaches often focus on specific MWE types, e.g., transformer-based models \citep{swaminathan2023, tanner2023} that incorporate linguistic features like dependency parsing for verbal discontinuous patterns \citep{nasr2015} and chunk features for nominal detection \citep{cordeiro2016}. Existing methods for all-type MWE rely on massive language models with billions of parameters \citep{ide2025} and lack systematic integration of linguistic knowledge. Moreover, while binary classification approaches have proven successful for span-based NER \citep{eberts2020, li2022t2ner, tan2020boundary, fei2021span}, to the best of our knowledge, this has not been explored yet for the task of MWE identification.

Several datasets support MWE research. PARSEME shared tasks \citep{savary2017, ramisch2018, ramisch2020, walsh2018} provide multilingual corpora for verbal MWEs. DiMSUM \citep{schneider2016} covers multiple types but suffers from annotation inconsistencies \citep{tanner2023}. STREUSLE \citep{schneider2018} augments the English Web Treebank with various MWE types and emphasizes supersense labeling over explicit MWE type distinctions. The recently introduced CoAM \citep{ide2025} addresses these limitations by offering comprehensive coverage of all MWE types with explicit type annotations (NOUN, VERB, MOD/CONN, CLAUSE, OTHER), enabling fine-grained analyses previously unavailable. While limited to 1,301 sentences, CoAM's careful curation provides crucial evaluation capabilities.

Leveraging CoAM's comprehensive annotations for primary evaluation, we present an approach that reformulates MWE identification as binary token-level classification—departing from traditional multi-class sequence labeling to predict three independent binary labels per token. Our DeBERTa-v3-large model \citep{he2021} achieves 69.8\% F1 on CoAM, surpassing the state-of-the-art on this dataset, Qwen-72B (57.8\% F1), by 12 points while using 165× fewer parameters. We further validate generalization on STREUSLE, achieving 78.9\% F1 after mapping its categories to CoAM-compatible types, confirming that our approach transfers effectively across datasets with different annotation schemes.

\section{Method}

We detail our approach, beginning with the dataset description and our binary token-level classification framework, followed by linguistic feature integration and data augmentation.

\begin{table*}[t]
\centering
\footnotesize
\setlength{\tabcolsep}{1.5pt}
\renewcommand{\arraystretch}{1.1}
\begin{tabular}{ll|ccccccccccccccccc}
\toprule
& & Q-72B & BBS & BBT & BLT & DBT & DLT & DBT+l & DLT+l & DBT+o & DLT+o & DBT+a & DLT+a & DBT+lo & \textbf{\underline{DLT+lo}} & DBT+la & DLT+la \\
\midrule
\multirow{3}{*}{\rotatebox{90}{\textbf{CoAM}}}
& \textbf{Prec} & \cellcolor{perf50}63.8 & \cellcolor{perf10}51.6 & \cellcolor{perf40}62.5 & \cellcolor{perf60}67.4 & \cellcolor{perf40}62.1 & \cellcolor{perf50}64.0 & \cellcolor{perf50}64.0 & \cellcolor{perf20}52.6 & \cellcolor{perf60}65.0 & \cellcolor{perf40}61.3 & \cellcolor{perf70}68.2 & \cellcolor{perf40}61.1 & \cellcolor{perf50}63.6 & \cellcolor{perf80}\textbf{69.3} & \cellcolor{perf60}66.4 & \cellcolor{perf90}70.0 \\
& \textbf{Rec} & \cellcolor{perf20}52.8 & \cellcolor{perf30}55.9 & \cellcolor{perf30}55.6 & \cellcolor{perf20}53.3 & \cellcolor{perf60}65.4 & \cellcolor{perf40}61.2 & \cellcolor{perf70}68.0 & \cellcolor{perf80}69.3 & \cellcolor{perf40}61.9 & \cellcolor{perf50}63.5 & \cellcolor{perf30}58.5 & \cellcolor{perf80}69.3 & \cellcolor{perf60}66.4 & \cellcolor{perf90}\textbf{70.3} & \cellcolor{perf50}63.8 & \cellcolor{perf60}66.4 \\
& \textbf{F1} & \cellcolor{perf30}57.8 & \cellcolor{perf20}53.7 & \cellcolor{perf30}58.9 & \cellcolor{perf30}59.5 & \cellcolor{perf50}63.7 & \cellcolor{perf40}62.6 & \cellcolor{perf60}65.9 & \cellcolor{perf30}59.8 & \cellcolor{perf50}63.4 & \cellcolor{perf40}62.4 & \cellcolor{perf40}63.0 & \cellcolor{perf50}64.9 & \cellcolor{perf50}65.0 & \cellcolor{perf90}\textbf{69.8} & \cellcolor{perf50}65.1 & \cellcolor{perf70}68.2 \\
\midrule
\multirow{3}{*}{\rotatebox{90}{\tiny\textbf{STREUSLE}}}
& \textbf{Prec} & -- & \cellcolor{perf0}11.1 & \cellcolor{perf30}58.1 & \cellcolor{perf50}68.4 & \cellcolor{perf50}68.8 & \cellcolor{perf40}64.2 & \cellcolor{perf60}71.3 & \cellcolor{perf70}75.5 & \cellcolor{perf50}67.1 & \cellcolor{perf40}62.1 & \cellcolor{perf50}68.4 & \cellcolor{perf40}63.3 & \cellcolor{perf60}71.5 & \cellcolor{perf60}72.4 & \cellcolor{perf80}\textbf{73.5} & \cellcolor{perf60}70.0 \\
& \textbf{Rec} & -- & \cellcolor{perf90}90.1 & \cellcolor{perf40}78.2 & \cellcolor{perf30}72.5 & \cellcolor{perf40}78.5 & \cellcolor{perf60}84.5 & \cellcolor{perf50}80.6 & \cellcolor{perf50}82.4 & \cellcolor{perf50}82.7 & \cellcolor{perf70}87.3 & \cellcolor{perf50}81.7 & \cellcolor{perf60}83.8 & \cellcolor{perf50}82.0 & \cellcolor{perf50}82.4 & \cellcolor{perf80}\textbf{85.2} & \cellcolor{perf60}84.5 \\
& \textbf{F1} & -- & \cellcolor{perf0}19.8 & \cellcolor{perf40}66.8 & \cellcolor{perf50}70.4 & \cellcolor{perf60}73.3 & \cellcolor{perf60}72.9 & \cellcolor{perf70}75.7 & \cellcolor{perf80}78.8 & \cellcolor{perf60}74.1 & \cellcolor{perf60}72.6 & \cellcolor{perf60}74.5 & \cellcolor{perf60}72.1 & \cellcolor{perf70}76.4 & \cellcolor{perf70}77.1 & \cellcolor{perf90}\textbf{78.9} & \cellcolor{perf70}76.5 \\
\bottomrule
\end{tabular}
\caption{Performance comparison on CoAM and STREUSLE test sets. On CoAM, DLT+lo achieves 69.8\% F1, outperforming Qwen-72B (57.8\% F1). On STREUSLE, DBT+la achieves 78.9\% F1. Best results per dataset in bold.}
\label{tab:main}
\end{table*}

\begin{table*}[!htbp]
\centering
\small
\setlength{\tabcolsep}{1.3pt}
\renewcommand{\arraystretch}{1.15}
\begin{tabular}{ll|ccccccccccccccccc}
\toprule
& & Q-72B & BBS & BBT & BLT & DBT & DLT & DBT+l & DLT+l & DBT+o & DLT+o & DBT+a & DLT+a & DBT+lo & DLT+lo & DBT+la & DLT+la \\
\midrule
\multirow{4}{*}{\rotatebox{90}{\textbf{CoAM}}}
& \textbf{CL} & \cellcolor{perf20}28.6 & \cellcolor{perf30}42.9 & \cellcolor{perf20}28.6 & \cellcolor{perf30}42.9 & \cellcolor{perf30}42.9 & \cellcolor{perf30}42.9 & \cellcolor{perf30}42.9 & \cellcolor{perf30}42.9 & \cellcolor{perf30}42.9 & \cellcolor{perf30}42.9 & \cellcolor{perf20}28.6 & \cellcolor{perf30}42.9 & \cellcolor{perf20}28.6 & \cellcolor{perf90}\textbf{85.7} & \cellcolor{perf30}42.9 & \cellcolor{perf70}71.4 \\
& \textbf{M/C} & \cellcolor{perf30}57.7 & \cellcolor{perf50}67.6 & \cellcolor{perf60}75.7 & \cellcolor{perf60}76.6 & \cellcolor{perf70}82.9 & \cellcolor{perf60}76.6 & \cellcolor{perf70}80.2 & \cellcolor{perf70}82.0 & \cellcolor{perf60}76.6 & \cellcolor{perf70}79.3 & \cellcolor{perf50}67.6 & \cellcolor{perf70}78.4 & \cellcolor{perf60}76.6 & \cellcolor{perf90}\textbf{84.7} & \cellcolor{perf60}73.0 & \cellcolor{perf70}80.2 \\
& \textbf{NOUN} & \cellcolor{perf20}42.1 & \cellcolor{perf30}43.8 & \cellcolor{perf10}38.0 & \cellcolor{perf10}35.5 & \cellcolor{perf40}57.9 & \cellcolor{perf20}51.2 & \cellcolor{perf50}60.3 & \cellcolor{perf60}65.3 & \cellcolor{perf30}53.7 & \cellcolor{perf30}52.9 & \cellcolor{perf30}53.7 & \cellcolor{perf40}62.7 & \cellcolor{perf50}60.3 & \cellcolor{perf70}\textbf{66.9} & \cellcolor{perf50}63.6 & \cellcolor{perf40}58.7 \\
& \textbf{VERB} & \cellcolor{perf40}59.7 & \cellcolor{perf40}59.0 & \cellcolor{perf30}56.8 & \cellcolor{perf20}51.1 & \cellcolor{perf40}59.7 & \cellcolor{perf40}59.0 & \cellcolor{perf60}\textbf{66.9} & \cellcolor{perf50}64.7 & \cellcolor{perf40}59.0 & \cellcolor{perf50}62.6 & \cellcolor{perf40}58.3 & \cellcolor{perf60}66.2 & \cellcolor{perf60}\textbf{66.9} & \cellcolor{perf60}\textbf{66.9} & \cellcolor{perf40}59.0 & \cellcolor{perf50}63.3 \\
\midrule
\multirow{3}{*}[-0.8em]{\rotatebox{90}{\tiny\textbf{STREUSLE}}}
& \textbf{M/C} & -- & \cellcolor{perf80}90.6 & \cellcolor{perf60}81.2 & \cellcolor{perf50}78.1 & \cellcolor{perf60}84.3 & \cellcolor{perf70}89.1 & \cellcolor{perf70}85.9 & \cellcolor{perf70}87.5 & \cellcolor{perf90}\textbf{92.2} & \cellcolor{perf80}90.6 & \cellcolor{perf70}89.1 & \cellcolor{perf70}89.1 & \cellcolor{perf70}87.5 & \cellcolor{perf60}84.4 & \cellcolor{perf90}\textbf{92.2} & \cellcolor{perf70}89.1 \\
& \textbf{NOUN} & -- & \cellcolor{perf70}88.4 & \cellcolor{perf50}76.9 & \cellcolor{perf40}69.2 & \cellcolor{perf50}78.5 & \cellcolor{perf70}86.1 & \cellcolor{perf60}83.1 & \cellcolor{perf60}84.6 & \cellcolor{perf60}82.3 & \cellcolor{perf70}\textbf{86.9} & \cellcolor{perf60}81.5 & \cellcolor{perf60}84.6 & \cellcolor{perf60}83.1 & \cellcolor{perf60}82.3 & \cellcolor{perf70}86.1 & \cellcolor{perf70}85.4 \\
& \textbf{VERB} & -- & \cellcolor{perf80}90.9 & \cellcolor{perf50}74.2 & \cellcolor{perf50}72.7 & \cellcolor{perf50}74.2 & \cellcolor{perf50}75.7 & \cellcolor{perf50}75.0 & \cellcolor{perf50}75.7 & \cellcolor{perf50}75.7 & \cellcolor{perf70}\textbf{81.8} & \cellcolor{perf50}77.2 & \cellcolor{perf50}75.7 & \cellcolor{perf50}75.7 & \cellcolor{perf60}80.3 & \cellcolor{perf60}78.9 & \cellcolor{perf60}80.3 \\
& \textbf{OTH} & -- & \cellcolor{perf90}95.8 & \cellcolor{perf70}87.5 & \cellcolor{perf50}75.0 & \cellcolor{perf50}75.0 & \cellcolor{perf70}87.5 & \cellcolor{perf50}72.7 & \cellcolor{perf50}75.0 & \cellcolor{perf60}79.2 & \cellcolor{perf90}\textbf{95.8} & \cellcolor{perf50}75.0 & \cellcolor{perf60}83.8 & \cellcolor{perf60}79.2 & \cellcolor{perf60}83.3 & \cellcolor{perf60}79.2 & \cellcolor{perf60}79.2 \\
\bottomrule
\end{tabular}
\caption{Type-specific recall (\%) across MWE types. CoAM: CL=CLAUSE (N=7), M/C=MOD/CONN (N=111), NOUN (N=121), VERB (N=139). STREUSLE: M/C (N=64), NOUN (N=130), VERB (N=66), OTH=OTHER (N=24). CoAM OTHER (N=3) excluded; STREUSLE has no CLAUSE annotations. Note: on STREUSLE, DLT+o achieves higher recall than DBT+la in 3/4 categories but lower overall F1 (72.6\% vs 78.9\%, see Table 1) due to over-prediction (399 vs 329 predictions against 284 gold MWEs).}
\label{tab:mwe_types}
\end{table*}

\begin{table*}[t]
\centering
\footnotesize
\setlength{\tabcolsep}{1.5pt}
\renewcommand{\arraystretch}{1.1}
\begin{tabular}{ll|ccccccccccccccccc}
\toprule
& & Q-72B & BBS & BBT & BLT & DBT & DLT & DBT+l & DLT+l & DBT+o & DLT+o & DBT+a & DLT+a & DBT+lo & \textbf{\underline{DLT+lo}} & DBT+la & DLT+la \\
\midrule
\multirow{6}{*}{\rotatebox{90}{\textbf{CoAM}}}
& \textbf{C-P} & -- & \cellcolor{perf10}51.6 & \cellcolor{perf70}77.4 & \cellcolor{perf80}\textbf{81.1} & \cellcolor{perf70}77.5 & \cellcolor{perf70}78.9 & \cellcolor{perf60}74.1 & \cellcolor{perf50}65.7 & \cellcolor{perf70}79.1 & \cellcolor{perf70}76.5 & \cellcolor{perf80}80.2 & \cellcolor{perf70}76.5 & \cellcolor{perf60}72.5 & \cellcolor{perf70}76.9 & \cellcolor{perf60}75.8 & \cellcolor{perf80}80.4 \\
& \textbf{C-R} & -- & \cellcolor{perf50}63.0 & \cellcolor{perf40}59.8 & \cellcolor{perf40}58.3 & \cellcolor{perf70}70.4 & \cellcolor{perf60}66.3 & \cellcolor{perf70}72.8 & \cellcolor{perf90}\textbf{74.9} & \cellcolor{perf60}66.0 & \cellcolor{perf60}67.5 & \cellcolor{perf50}62.1 & \cellcolor{perf80}73.1 & \cellcolor{perf70}70.1 & \cellcolor{perf90}\textbf{74.9} & \cellcolor{perf60}67.8 & \cellcolor{perf70}70.4 \\
& \textbf{C-F1} & \cellcolor{perf30}57.3 & \cellcolor{perf30}56.6 & \cellcolor{perf60}67.4 & \cellcolor{perf60}67.8 & \cellcolor{perf80}73.8 & \cellcolor{perf70}72.0 & \cellcolor{perf80}73.4 & \cellcolor{perf70}70.0 & \cellcolor{perf70}71.9 & \cellcolor{perf70}71.7 & \cellcolor{perf70}70.0 & \cellcolor{perf80}74.7 & \cellcolor{perf70}71.3 & \cellcolor{perf90}\textbf{75.9} & \cellcolor{perf70}71.6 & \cellcolor{perf80}75.1 \\
& \textbf{D-P} & -- & \cellcolor{perf0}0.0 & \cellcolor{perf10}12.8 & \cellcolor{perf10}10.3 & \cellcolor{perf10}11.7 & \cellcolor{perf10}11.3 & \cellcolor{perf10}17.8 & \cellcolor{perf10}9.4 & \cellcolor{perf10}16.0 & \cellcolor{perf10}14.4 & \cellcolor{perf20}20.0 & \cellcolor{perf10}15.6 & \cellcolor{perf20}22.5 & \cellcolor{perf30}\textbf{25.9} & \cellcolor{perf20}21.9 & \cellcolor{perf20}23.1 \\
& \textbf{D-R} & -- & \cellcolor{perf0}0.0 & \cellcolor{perf20}23.3 & \cellcolor{perf10}14.0 & \cellcolor{perf20}25.6 & \cellcolor{perf20}20.9 & \cellcolor{perf30}30.2 & \cellcolor{perf20}25.6 & \cellcolor{perf30}30.2 & \cellcolor{perf30}32.6 & \cellcolor{perf30}30.2 & \cellcolor{perf40}\textbf{39.5} & \cellcolor{perf40}37.2 & \cellcolor{perf30}34.9 & \cellcolor{perf30}32.6 & \cellcolor{perf30}34.9 \\
& \textbf{D-F1} & \cellcolor{perf10}17.1 & \cellcolor{perf0}0.0 & \cellcolor{perf10}16.5 & \cellcolor{perf10}11.9 & \cellcolor{perf10}16.1 & \cellcolor{perf10}14.6 & \cellcolor{perf20}22.4 & \cellcolor{perf10}13.8 & \cellcolor{perf20}21.0 & \cellcolor{perf20}20.0 & \cellcolor{perf20}24.1 & \cellcolor{perf20}22.4 & \cellcolor{perf30}28.1 & \cellcolor{perf30}\textbf{29.7} & \cellcolor{perf20}26.2 & \cellcolor{perf30}27.8 \\
\midrule
\multirow{3}{*}[-2em]{\rotatebox{90}{\tiny\textbf{STREUSLE}}}
& \textbf{C-P} & -- & \cellcolor{perf10}29.5 & \cellcolor{perf60}78.3 & \cellcolor{perf70}84.2 & \cellcolor{perf70}85.7 & \cellcolor{perf70}84.3 & \cellcolor{perf70}84.4 & \cellcolor{perf80}89.4 & \cellcolor{perf80}87.6 & \cellcolor{perf70}84.1 & \cellcolor{perf80}\textbf{89.3} & \cellcolor{perf70}83.6 & \cellcolor{perf70}86.2 & \cellcolor{perf70}85.7 & \cellcolor{perf80}88.3 & \cellcolor{perf70}84.7 \\
& \textbf{C-R} & -- & \cellcolor{perf80}91.8 & \cellcolor{perf60}82.3 & \cellcolor{perf50}75.6 & \cellcolor{perf60}82.3 & \cellcolor{perf80}88.6 & \cellcolor{perf70}85.1 & \cellcolor{perf70}86.3 & \cellcolor{perf70}86.2 & \cellcolor{perf80}91.3 & \cellcolor{perf70}85.5 & \cellcolor{perf80}87.8 & \cellcolor{perf70}85.5 & \cellcolor{perf70}85.1 & \cellcolor{perf80}\textbf{88.6} & \cellcolor{perf80}87.4 \\
& \textbf{C-F1} & -- & \cellcolor{perf20}44.6 & \cellcolor{perf60}80.3 & \cellcolor{perf60}79.7 & \cellcolor{perf70}84.0 & \cellcolor{perf80}86.4 & \cellcolor{perf70}84.7 & \cellcolor{perf80}87.8 & \cellcolor{perf80}86.9 & \cellcolor{perf80}87.6 & \cellcolor{perf80}87.4 & \cellcolor{perf70}85.6 & \cellcolor{perf70}85.8 & \cellcolor{perf70}85.4 & \cellcolor{perf90}\textbf{88.4} & \cellcolor{perf80}86.1 \\
& \textbf{D-P} & -- & \cellcolor{perf0}1.4 & \cellcolor{perf10}10.5 & \cellcolor{perf10}18.0 & \cellcolor{perf10}16.4 & \cellcolor{perf10}13.2 & \cellcolor{perf10}18.7 & \cellcolor{perf20}21.9 & \cellcolor{perf10}15.1 & \cellcolor{perf10}12.3 & \cellcolor{perf10}14.7 & \cellcolor{perf10}13.0 & \cellcolor{perf20}20.5 & \cellcolor{perf20}21.9 & \cellcolor{perf30}\textbf{24.3} & \cellcolor{perf20}21.2 \\
& \textbf{D-R} & -- & \cellcolor{perf60}75.8 & \cellcolor{perf30}41.3 & \cellcolor{perf30}44.8 & \cellcolor{perf30}44.8 & \cellcolor{perf30}48.3 & \cellcolor{perf30}41.4 & \cellcolor{perf30}48.3 & \cellcolor{perf40}51.7 & \cellcolor{perf40}51.7 & \cellcolor{perf30}48.3 & \cellcolor{perf30}48.3 & \cellcolor{perf40}51.7 & \cellcolor{perf40}55.2 & \cellcolor{perf50}\textbf{58.6} & \cellcolor{perf50}\textbf{58.6} \\
& \textbf{D-F1} & -- & \cellcolor{perf0}2.8 & \cellcolor{perf10}16.7 & \cellcolor{perf20}25.7 & \cellcolor{perf20}24.1 & \cellcolor{perf20}20.7 & \cellcolor{perf20}25.8 & \cellcolor{perf30}30.1 & \cellcolor{perf20}23.4 & \cellcolor{perf20}19.9 & \cellcolor{perf20}22.6 & \cellcolor{perf20}20.4 & \cellcolor{perf30}29.4 & \cellcolor{perf30}31.3 & \cellcolor{perf40}\textbf{34.4} & \cellcolor{perf30}31.2 \\
\bottomrule
\end{tabular}
\caption{Performance on continuous (C) and discontinuous (D) MWEs. CoAM: N=338 continuous, N=43 discontinuous. STREUSLE: N=255 continuous, N=29 discontinuous. Best model per dataset in bold.}
\label{tab:continuity}
\end{table*}

\subsection{Datasets}
\textbf{CoAM} \citep{ide2025} contains 1,301 English sentences (780 train, 521 test) with 867 annotated MWEs, distributed over the following types: NOUN (34.3\% compounds and idiomatic nominals), VERB (37.5\% phrasal and light verb constructions), MOD/CONN (23.9\% complex modifiers), CLAUSE (1.6\% sentential idioms), and OTHER (2.8\%). MWEs are 88.7\% continuous and 11.3\% discontinuous, spanning 2--13 tokens.

\textbf{STREUSLE} \citep{schneider2018} v4.6 (UD v2.16) provides a complementary evaluation setting with 1,530 training (2,448 MWEs), 217 development (287 MWEs), and 209 test (284 MWEs) sentences. We map STREUSLE's fine-grained categories (e.g., \textit{verb-particle/construction}, \textit{noun/compound}) to CoAM-compatible types: verbal constructions to VERB, nominal expressions to NOUN, adverbial/prepositional phrases to MOD/CONN, and remaining categories to OTHER. No STREUSLE annotations correspond to the CLAUSE category. MWEs span 2--7 tokens, with 10.2\% being discontinuous.

\subsection{Binary Token-Level Classification and MWE Reconstruction}
We reformulate MWE identification as binary token-level classification, inspired by its success in NER \citep{eberts2020, li2022t2ner, tan2020boundary, fei2021span}: instead of assigning each token a single label from a multi-class tagset (e.g., B-MWE, I-MWE, O), we make three independent binary decisions per token, predicting whether it serves as a START, END, or INSIDE component of a MWE. Our approach predicts three independent probabilities for each token $x_i$: $p_i^{\text{start}}$, $p_i^{\text{end}}$, and $p_i^{\text{inside}}$. For instance, in \textit{looked the information up}:
\begin{itemize}
\item \textit{looked}: START=1, END=0, INSIDE=0
\item \textit{the}: START=0, END=0, INSIDE=0
\item \textit{information}: START=0, END=0, INSIDE=0
\item \textit{up}: START=0, END=1, INSIDE=0
\end{itemize}
Our model would ideally predict high START probability for \textit{looked}, high END probability for \textit{up}, and low INSIDE probabilities for the intervening words, correctly identifying the discontinuous MWE \{\textit{looked}, \textit{up}\}. Crucially, our binary token-level approach scales linearly with sequence length $O(n)$, as we make three independent predictions per token, whereas span-based enumeration requires quadratic complexity $O(n^2)$ to evaluate all possible token pairs.

To train this binary classification framework, we must first convert CoAM's original span-based annotations. CoAM provides span-based annotations (token indices for each MWE), which we convert to token-level labels through a projection system generating three binary vectors that form our START/END/INSIDE encoding scheme:
$\text{start}[i] = 1$ if token $i$ is the first token of any MWE,
$\text{end}[i] = 1$ if token $i$ is the last token of any MWE, and
$\text{inside}[i] = 1$ if token $i$ appears strictly between start/end positions.
Projections are stored as versioned JSON artifacts with SHA256 hashing for reproducibility.

At inference, these independent predictions must be reconstructed into MWE spans. We form candidate spans from token pairs $(s,e)$ with $e>s$ such that $p^{\text{start}}_s \geq \tau_{\text{start}}$ and $p^{\text{end}}_e \geq \tau_{\text{end}}$. A span is accepted if its width $(e-s+1) \leq 13$ (dataset maximum), intermediate tokens $t \in (s,e)$ are included when $p^{\text{inside}}_t \geq \tau_{\text{inside}}$, and the reconstructed expression contains 2–6 members, matching CoAM's MWE length distribution. Candidates are stored as sorted sets of token indices. After training, we tune the three thresholds ($\tau_{\text{start}}$, $\tau_{\text{end}}$, $\tau_{\text{inside}}$) independently on the development set via grid search, finding optimal values that maximize the F1 score. Our approach thus focuses on MWE segmentation and boundary detection; type classification could be added as a second-stage task but is beyond the scope of this work (see Appendix A).

\subsection{Linguistic Feature Integration}
We enrich DeBERTa-v3-large with two forms of syntactic knowledge:

\textbf{NP Chunk Features.} Tokens are tagged as inside or outside a noun phrase (NP) using spaCy en\_core\_web\_lg \citep{honnibal2020}. These binary tags are mapped to 16-dimensional learnable embeddings and concatenated with contextualized token states. Chunk embeddings help the model detect nominal MWEs such as \textit{stock market}, a class noted in the CoAM study as particularly difficult to recall compared to MOD/CONN or VERB MWEs.

\textbf{Dependency features.} Following extensive prior work on dependency-based MWE identification \citep{lin1999, wehrli2000, green2013, nasr2015}, we compute shortest dependency path lengths between all token pairs using NetworkX \citep{hagberg2008} graph traversal, capped at 5. These distances serve dual purposes: (1) as learnable embeddings during encoding, where we average each token's dependency distances to other tokens and map them to 32-dimensional embeddings, and (2) as hard constraints during reconstruction, where we reject discontinuous MWE candidates with consecutive members having dependency distance greater than 4, filtering out implausible token combinations.

\subsection{Data Augmentation}
Beyond linguistic features, addressing the severe class imbalance in the training data proved crucial for model performance. We evaluated two data augmentation approaches \citep{he2009}: (1) oversampling of training sentences containing MWEs, and (2) lexical substitution, replacing single tokens outside MWE boundaries with semantically similar words. Both approaches were tested by selecting 10\%, 20\%, 30\%, or 40\% of the training sentences for augmentation.

The effectiveness of each strategy depends on dataset size. On the smaller CoAM (486 training MWEs), oversampling proved more effective, with 30\% selection ratio proving optimal. We attribute this to CoAM's limited size combined with its carefully curated annotations: the dataset appears too small for the model to generalize from lexical variations, while any modifications risk disrupting the subtle patterns that define MWEs. Conversely, on the larger STREUSLE (2,448 training MWEs), lexical substitution proved more effective, as sufficient training examples enable the model to benefit from lexical variation.

\section{Experiments}

We conduct a systematic ablation study with 15 model configurations to isolate the contributions of each component: our binary token-level classification approach, model architecture choice, linguistic feature integration, and data augmentation.

\subsection{Experimental Design}
Our experiments build incrementally across four phases. First, we compare BERT-base span-based prediction (BBS) against BERT-base/large token-level classification (BBT/BLT). Second, we compare BBT/BLT against DeBERTa-v3-base/large (DBT, DLT), whose disentangled attention should better capture discontinuous patterns \citep{he2021}. Third, we compare DBT/DLT against variants augmented with linguistic features (+l) targeting NOUN and discontinuous MWEs. Finally, we compare DBT+l/DLT+l against data-augmented variants using oversampling (+o) and lexical substitution (+a), both independently and combined with linguistic features (see Table 1).

\subsection{Setup}
We convert the span-based annotations from both datasets to our START/END/INSIDE format by projecting the provided token indices. For STREUSLE, we additionally apply the type mapping described in Section 2.1. This harmonization enables direct comparison across datasets while preserving type-specific analysis capabilities.

We train all models on NVIDIA GH200 hardware. Our baseline for CoAM is the dataset authors' Qwen-72B (57.8\% F1, \citealp{ide2025}), and BERT-base span-based (BBS) as additional comparison across both datasets. We report exact span matching micro F1 with breakdowns by type and continuity. Hyperparameters were optimized via Optuna across learning rate, batch size, dropout, and weight decay. Our best model on CoAM (DLT+lo) uses learning rate 3e-5, batch size 16, dropout 0.3, weight decay 0.01 with early stopping at epoch 8. Our best model on STREUSLE (DBT+la) uses learning rate 2e-5, batch size 4, dropout 0.3, weight decay 0.05 with early stopping at epoch 10. Training requires 10--25 minutes per model, making our approach highly practical compared to large model fine-tuning.

\subsection{Main Results}

Our systematic ablation results in Table~\ref{tab:main} reveal several critical insights about each component's contribution. Our binary token-level classification yields +5.2 F1 improvement over span-based prediction (BBS: 53.7 → BBT: 58.9), validating that three independent START/END/INSIDE decisions per token provide richer gradient signals than sparse span-level feedback. DBT achieves 63.7\% F1, surpassing BBT by +4.8 points and even outperforming BLT, confirming that disentangled attention mechanisms are particularly suited for MWE detection where content and position interact in an intricate way.

Adding linguistic features to DBT improves F1 by +2.2 points (63.7 → 65.9), with particularly strong recall improvements. However, DLT shows a surprising decrease with linguistic features alone (62.6 → 59.8), suggesting potential overfitting on the small dataset. A critical pattern emerges in the interaction between augmentation and model capacity: without augmentation, base models often outperform large counterparts, but with augmentation, large models excel—DLT+lo achieves 69.8\% F1 while DBT+lo reaches only 65.0\%. Oversampling consistently outperforms lexical substitution when combined with linguistic features, which may be attributed to CoAM's high annotation quality where lexical changes risk disrupting subtle MWE patterns (see Appendix A.2 and Appendix B).

Our DeBERTa-v3-large model achieves 69.8\% F1 on CoAM, surpassing the state-of-the-art on this dataset, Qwen-72B (57.8\% F1), by 12 points while using 165× fewer parameters. Other comparisons to previous work are complex because we focus on all types of MWEs, which differs fundamentally from recent frameworks like \citet{swaminathan2023} that are tailored specifically to verbal MWEs or binary idiomaticity classification rather than comprehensive span identification across all MWE types (see Framework Comparison Limitations).

On STREUSLE, the ablation pattern replicates. The baseline BBS achieves 19.8\% F1 due to massive over-prediction (11.1\% precision), while binary token-level classification provides substantial improvement: DBT reaches 73.3\% F1, a 53.5-point gain mirroring CoAM's pattern. Linguistic features consistently improve both backbones, with DLT+l achieving 78.8\% F1. Crucially, the optimal augmentation strategy differs: lexical substitution outperforms oversampling (DBT+la: 78.9\% F1 vs DBT+lo: 76.4\% F1), contrasting with CoAM where oversampling proved superior---we attribute this to STREUSLE's larger training set (2,448 MWEs) enabling the model to benefit from lexical variation.

\subsection{Analysis by MWE Type and Continuity}

Table~\ref{tab:continuity}\footnote{Type-specific precision cannot be computed as our model predicts only reconstructed span boundaries without type labels. See Limitations (Type Prediction Limitations) for a detailed discussion.} highlights dramatic improvements for discontinuous MWEs. The baseline span-based model completely fails on discontinuous MWEs (0\% recall), while our binary token-level classification immediately enables detection (BBT: 23.3\% recall). Adding linguistic features to DBT increases discontinuous recall by +4.6 points, confirming that explicit syntactic knowledge helps identify separated MWE components. Our best model on CoAM (DLT+lo) achieves 34.9\% discontinuous recall while maintaining 25.9\% precision (see Appendix B for detailed error analysis).

In addition to handling discontinuities, we also analyze performance across MWE types to understand where improvements are most pronounced. Table~\ref{tab:mwe_types} demonstrates type-specific improvements across categories. CLAUSE expressions improve from Qwen-72B's 28.6\% to 85.7\% recall, though with only 7 test instances, this represents detecting 6 versus 2 examples. NP chunking knowledge integration helps NOUN expressions, improving recall from 57.9\% to 60.3\% (DBT → DBT+l), with stronger effects in large models. MOD/CONN expressions show consistent gains, with DLT+lo achieving 84.7\% recall compared to Qwen's 57.7\%. Importantly, type-specific recall must be interpreted alongside precision: models with aggressive prediction strategies can achieve high recall across categories while suffering from low overall F1 due to excessive false positives.

On STREUSLE, type-specific patterns confirm generalization: MOD/CONN achieves 92.2\% recall for DBT+la, followed by NOUN (86.1\%) and VERB (78.9\%). Regarding contiguity patterns, DBT+la achieves 88.4\% F1 on continuous MWEs and 34.4\% F1 on discontinuous ones. Linguistic features consistently boost discontinuous performance: DLT+l reaches 30.1\% discontinuous F1 versus DLT's 20.7\%, confirming that syntactic structure helps across datasets.

\section{Conclusion}
We reformulate MWE identification as binary token-level classification with three independent START/END/INSIDE predictions per token, combining DeBERTa-v3 with NP chunking and dependency features plus data augmentation. On CoAM, this achieves 69.8\% F1, outperforming Qwen-72B (57.8\%) by 12 points with 165× fewer parameters; for discontinuous MWEs, we achieve 29.7\% F1 versus Qwen-72B's 17.1\%, doubling baseline performance though these patterns remain challenging. We confirm generalization on STREUSLE, achieving 78.9\% F1 with the same methodological choices, demonstrating that binary token-level classification and linguistic feature integration capture fundamental aspects of MWE structure rather than dataset-specific artifacts.

\label{sec:bibtex}

\section*{Limitations}

While our approach achieves substantial improvements on CoAM, several limitations warrant discussion:

\textbf{Dataset Scale and Model Generalization.} The first dataset we used, CoAM, is of high quality and diverse but it is small. This may lead to potential overfitting (see Appendix A.2 for detailed analysis), but meaningful generalization is evident: the test set remains unseen during training and is guaranteed to not contain MWE patterns present in the training data, yet our model maintains strong performance. The performance difference between training and test sets (Table~\ref{tab:train_test_gap}) is typical of small datasets. 

On the second dataset, STREUSLE, which is larger in its training set, we confirm cross-dataset transfer (78.9\% F1). The finding that simple duplication outperforms lexical substitution on CoAM while the opposite holds for STREUSLE (Section 2.4) suggests that dataset size influences optimal augmentation strategy, though preserving exact MWE patterns remains crucial as even minor lexical variations can fundamentally alter MWEs' idiomatic properties.

\textbf{Statistical Robustness.} Preliminary experiments with different random seeds on a subset of models (5 out of 15 configurations) showed no significant performance variation from the reported results using seed 42, suggesting stability in our approach. However, due to limited time for resource access, we could not systematically test all model configurations across multiple seeds.

\textbf{Framework Comparison Limitations.} Recent work by \citet{swaminathan2023} achieves strong results on MWE-related tasks but addresses fundamentally different problems than comprehensive span detection, preventing direct comparison. Their MTLB-STRUCT framework targets PARSEME 1.2, which exclusively annotates verbal MWEs (light-verb constructions, verb-particle constructions, inherently reflexive verbs, and verbal idioms), making it architecturally incompatible with CoAM where 62.5\% of MWEs are non-verbal (NOUN, MOD/CONN, CLAUSE, OTHER categories). Their second approach addresses SemEval 2022 Task 2, a binary sentence-level classification task determining whether a known potentially-idiomatic expression is used idiomatically or literally—fundamentally different from identifying which token spans constitute MWEs.

\textbf{Discontinuous MWE Performance.} Despite doubling the baseline's recall for discontinuous MWE identification (34.9\% vs 17.1\%), the overall discontinuous F1 remains modest at 29.7\%. The dependency distance constraints we introduce (rejecting candidates with consecutive members having distance >4) improve precision but may be overly restrictive for certain discontinuous patterns. This fundamental challenge suggests that even with linguistic enhancements, current architectures struggle with long-range discontinuous dependencies.

\textbf{Type Prediction Limitations.} Our binary token-level formulation predicts only span boundaries without MWE type labels, creating fundamental constraints on type-specific evaluation. While we can compute type-specific recall by checking which gold MWEs of each type were correctly identified, type-specific precision is impossible to calculate. When the model predicts a span like \textit{systemic risk} as an MWE (false positive), we have no type prediction to determine whether the model believes this is a NOUN, VERB, or MOD/CONN expression—we only know it identified these tokens as forming some MWE. Without type-specific precision, F1 scores per type cannot be computed. This limitation means Table 2 reports only recall metrics, preventing complete type-wise performance analysis. For instance, while we know the model correctly identifies 66.9\% of VERB expressions (recall), we cannot determine what percentage of its verb-like predictions are actually correct (precision) because the predictions lack type labels. This design choice prioritizes boundary detection accuracy over categorization but limits practical applications requiring typed MWE identification and restricts our ability to analyze whether false positives tend to be specific types of spurious expressions. Note that this issue only arises in the type-specific evaluation results in Table 2.

\textbf{Knowledge Source Dependencies.} Our linguistic features rely on spaCy's accuracy for NP chunking and dependency parsing. Errors in these upstream components propagate through our pipeline: incorrect chunk boundaries or erroneous dependency trees directly impact model performance. Additionally, such models and pipeline components (parsers, taggers) may not be available for all languages.

\textbf{Computational Requirements.} Experiments were conducted on NVIDIA GH200 hardware, which may limit reproducibility for researchers without access to high-end GPUs. While training requires only 10-25 minutes per model, the hyperparameter search with multiple trials across multiple oversampling ratios demands substantial compute. Our fastest configuration still requires $\sim$4GB GPU memory, potentially excluding older hardware.

\textbf{Language Coverage.} Evaluation is restricted to English, despite MWEs being a pervasive multilingual phenomenon with language-specific characteristics. The syntactic features we exploit (NP chunks, dependency paths) may transfer poorly to languages with different syntactic structures or morphological complexity.

\section*{Acknowledgments}
This research was funded by the NCCR Evolving Language, Swiss National Science Foundation Agreement No. 51NF40\_180888. We thank the anonymous reviewers for their helpful feedback. We also thank the people participating in the workshop \textit{Linguistic Representations: Through the lens of compounding}, organized by Prajit Dhar, for the fruitful discussions.

\bibliography{custom}

\clearpage
\appendix
\section{Implementation and Performance Details}

\begin{figure*}[!ht]
\centering
\includegraphics[width=0.9\textwidth]{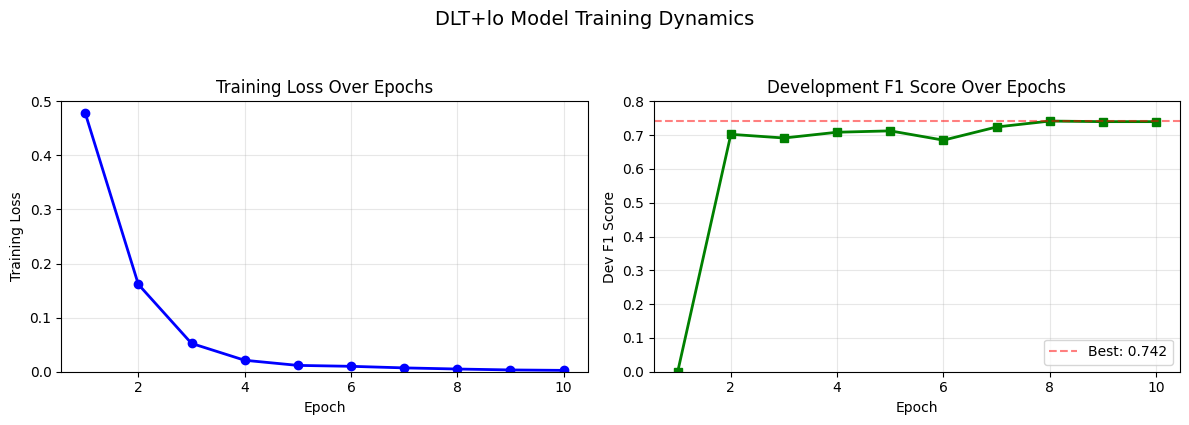}
\caption{Learning curves for DLT+lo: training loss decreases while development F1 plateaus at epoch 8 (approximately 0.74), triggering early stopping. The initial fluctuation (epochs 2-3) reflects the model adapting to the challenging task of distinguishing genuine MWEs from compositional phrases.}
\label{fig:learning_curves}
\end{figure*}

\subsection{Threshold Values and Hyperparameters}

Table~\ref{tab:thresholds} presents the optimal threshold values obtained through grid search on the development set for each model configuration. The thresholds control the sensitivity of START, END, and INSIDE predictions during MWE reconstruction. Lower thresholds generally increase recall at the cost of precision.

\begin{table}[!ht]
\centering
\begin{tabular}{l@{\hspace{1cm}}c@{\hspace{1cm}}c@{\hspace{1cm}}c}
\toprule
\textbf{Model} & \textbf{$\tau_{\text{start}}$} & \textbf{$\tau_{\text{end}}$} & \textbf{$\tau_{\text{inside}}$} \\
\midrule
\multicolumn{4}{l}{\textit{CoAM}} \\
BBS & -- & -- & -- \\
BBT & 0.45 & 0.20 & 0.20 \\
BLT & 0.60 & 0.20 & 0.20 \\
DBT & 0.45 & 0.60 & 0.20 \\
DLT & 0.50 & 0.20 & 0.20 \\
DBT+l & 0.30 & 0.20 & 0.20 \\
DLT+l & 0.50 & 0.30 & 0.20 \\
DBT+o & 0.20 & 0.20 & 0.20 \\
DLT+o & 0.50 & 0.60 & 0.40 \\
DBT+a & 0.50 & 0.60 & 0.30 \\
DLT+a & 0.40 & 0.40 & 0.20 \\
DBT+lo & 0.30 & 0.50 & 0.40 \\
\textbf{DLT+lo} & \textbf{0.50} & \textbf{0.60} & \textbf{0.20} \\
DBT+la & 0.60 & 0.40 & 0.20 \\
DLT+la & 0.60 & 0.40 & 0.20 \\
\midrule
\multicolumn{4}{l}{\textit{STREUSLE}} \\
\textbf{DBT+la} & \textbf{0.60} & \textbf{0.40} & \textbf{0.20} \\
\bottomrule
\end{tabular}
\caption{Optimal threshold values for each model configuration. BBS uses span-based prediction and does not require thresholds. Best models per dataset shown in bold.}
\label{tab:thresholds}
\end{table}

All models were trained with AdamW optimizer and linear learning rate scheduling. The best performing model (DLT+lo) used: learning rate 3e-5, batch size 16, dropout 0.3, weight decay 0.01, and early stopping with patience 3.

\subsection{Training Dynamics and Model Performance}

Table~\ref{tab:train_test_gap} shows training and test performance across all configurations. The train-test gaps (17.0-30.7\% F1) are expected given the limited training data (780 sentences) and the complexity of distinguishing idiomatic from compositional expressions. Notably, our best model maintains 69.8\% test F1 on unseen data—the test set contains MWE patterns and combinations not present in the training data—validating our architectural choices despite the challenging dataset size.

\begin{table}[!ht]
\centering
\small
\begin{tabular}{lccc}
\toprule
\textbf{Model} & \textbf{Train F1 (\%)} & \textbf{Test F1 (\%)} & \textbf{Gap ($\Delta$)} \\
\midrule
\multicolumn{4}{l}{\textit{CoAM}} \\
BBS & 70.7 & 53.7 & 17.0 \\
BBT & 89.6 & 58.9 & 30.7 \\
BLT & 90.1 & 59.5 & 30.6 \\
DBT & 83.2 & 63.7 & 19.5 \\
DLT & 88.9 & 62.6 & 26.3 \\
DBT+l & 92.3 & 65.9 & 26.4 \\
DLT+l & 86.1 & 59.8 & 26.3 \\
DBT+o & 89.3 & 63.4 & 25.9 \\
DLT+o & 88.7 & 62.4 & 26.3 \\
DBT+a & 86.5 & 63.0 & 23.5 \\
DLT+a & 89.6 & 64.9 & 24.7 \\
DBT+lo & 94.1 & 65.0 & 29.1 \\
\textbf{DLT+lo} & \textbf{95.0} & \textbf{69.8} & \textbf{25.2} \\
DBT+la & 93.9 & 65.1 & 28.8 \\
DLT+la & 94.6 & 68.2 & 26.4 \\
\midrule
\multicolumn{4}{l}{\textit{STREUSLE}} \\
\textbf{DBT+la} & \textbf{86.8} & \textbf{78.9} & \textbf{7.9} \\
\bottomrule
\end{tabular}
\caption{Train vs test F1 scores across all configurations. Best models per dataset shown in bold. STREUSLE's smaller gap reflects its larger training set.}
\label{tab:train_test_gap}
\end{table}

Figure~\ref{fig:learning_curves} illustrates the training dynamics of our best model. The plateauing development performance after epoch 5, while training loss continues decreasing, demonstrates effective early stopping preventing overtraining.

\section{Error Analysis}
\label{sec:appendix_b}

This appendix analyzes our best model's (DLT+lo) predictions on the test set to understand both its strong performance (69.8\% F1) and areas for improvement. The analysis reveals systematic patterns in the model's decisions, demonstrating that it has learned meaningful linguistic regularities rather than arbitrary memorization.

\subsection{Overall Performance Patterns}

The model generates 387 predictions for 381 gold MWEs, achieving well-balanced precision (69.3\%) and recall (70.3\%). Notably, 359 of 521 gold sentences (68.9\%) are predicted perfectly, indicating robust pattern learning. Among the 162 imperfect sentences out of 521 gold ones, 87 contain only false negatives (missed MWEs), 51 contain only false positives (spurious detections), and 24 show both error types, typically representing partial recognition where the model correctly identifies MWE presence but misaligns boundaries.

The prediction distribution reveals interesting patterns: 329 continuous and 58 discontinuous predictions against 338 continuous and 43 discontinuous gold MWEs. While the continuous predictions aligns well, discontinuous predictions show higher variance, with 43 false positives (36.1\% of all FPs) despite discontinuous MWEs comprising only 11.3\% of the gold standard.

\subsection{Continuous MWE Recognition}

Continuous expressions achieve strong 74.9\% recall (253/338 correctly identified). This 74.9\% recall represents a significant achievement, correctly identifying three-quarters of continuous MWEs despite training on only 780 sentences. Performance varies systematically by expression type:

\textbf{High-confidence categories}: Discourse markers (\textit{In fact}, \textit{Of course}, \textit{According to}) and phrasal verbs (\textit{took over}, \textit{brought down}, \textit{turn out}) achieve near-perfect detection when components are adjacent. Fixed nominals like \textit{Prime Minister} (correctly identified in all test instances) and \textit{exchange rate} demonstrate reliable institutional term recognition.

\textbf{False positives (76 instances)}: These split into interpretable categories. Technical overdetection (23 cases) flags domain bigrams like \textit{systemic risk} as MWEs--- specialized usage of this bigram with high co-occurrence, but arguably not a MWE. Incomplete spans (19 cases) systematically miss required elements, such as predicting \textit{open the way} instead of the gold \textit{open the way for}.

\textbf{False negatives (85 instances)}: The model fails to detect MWEs that appear infrequently in training (<7 occurrences). These include: (1) metaphorical phrases like \textit{watershed moment} and \textit{rose-tinted glasses} that require semantic understanding beyond surface patterns, and (2) short functional MWEs like \textit{act on} and \textit{yet again}, where the individual words (e.g., "act", "on") appear frequently throughout the corpus in non-MWE contexts, making it difficult for the model to recognize when they specifically form a MWE.

\subsection{Discontinuous MWE Progress}

Discontinuous detection demonstrates meaningful progress with 34.9\% recall (15/43), doubling baseline performance and achieving 25.9\% precision. 

\textbf{Successful patterns}: The 15 correct predictions follow learnable structures: (1) verb-particle with single interveners like \textit{turn [the situation] into} where dependency distance $\le 3$, (2) possessive frames like \textit{on [their] own}, and (3) simple modifier insertions like \textit{free [X] markets}.

\textbf{Error analysis}: The 43 discontinuous false positives primarily result from independent START/END predictions creating semantically invalid spans (\textit{not...risk} spanning 8 tokens) or partial captures missing critical elements (predicting \textit{turning...on} instead of the gold \textit{turning...gun...on}). The 28 false negatives often exceed our dependency distance threshold, like \textit{called...back} with an 11-token gap, or involve multiple discontinuities our architecture cannot represent.

\subsection{Type-Specific Performance}
\begin{table}[h]
\centering
\footnotesize
\begin{tabular}{lcccc}
\toprule
\textbf{Type} & \textbf{Rec\%} &\textbf{N} \\
\midrule
CLAUSE & 85.7 & 7 \\
MOD/CONN & 84.7 & 111 \\
VERB & 66.9 & 139 \\
NOUN & 66.9 & 121 \\
\bottomrule
\end{tabular}
\caption{Type-specific recall performance across MWE categories for our best model (DLT+lo).}
\label{tab:type_perf}
\end{table}

The model demonstrates strong performance across different MWE types. Formulaic expressions (MOD/CONN) achieve excellent 84.7\% recall, successfully identifying phrases like \textit{In fact} and \textit{According to} that follow consistent syntactic patterns. CLAUSE expressions reach 85.7\% recall despite having only 7 test instances, demonstrating the model's ability to generalize from limited examples. 

VERB and NOUN categories both achieve 66.9\% recall, a notable accomplishment given their inherent challenges. VERB expressions require distinguishing idiomatic uses (e.g., \textit{take off} meaning "succeed") from compositional ones (physically removing something). NOUN expressions span from technical terms to creative compounds, requiring the model to recognize both conventional phrases (\textit{stock market}) and novel compositional combinations.

Overall, the error analysis reveals that our model's mistakes are systematic and interpretable, often reflecting genuine linguistic ambiguities (e.g., compositional vs. idiomatic uses) or structural challenges rather than random failures.

\end{document}